\documentclass[twocolumn]{svjour3} % twocolumn
\smartqed  % flush right qed marks, e.g. at end of proof

\makeatletter
\renewcommand{\paragraph}{%
  \@startsection{paragraph}{4}%
  {\z@}{1ex \@plus 1ex \@minus .2ex}{-1em}%
  {\normalfont\normalsize\bfseries}%
}

\makeatother

\usepackage{graphicx}
\usepackage{times}
\usepackage{epsfig}
\usepackage{graphicx}
\usepackage{amsmath}
\usepackage{amssymb}
\usepackage{multicol}
\usepackage{balance}
\usepackage{multirow}
\usepackage{caption}
\usepackage{ctable}
\usepackage{array}
\usepackage{makecell}
\usepackage{subfig}
\usepackage{tabularx}
\usepackage{bbding}
\usepackage{pifont}
\usepackage{xfrac}
\usepackage{tabto,enumitem}

\usepackage{xspace}
\newcommand*{\eg}{e.g.\@\xspace}
\newcommand*{\ie}{i.e.\@\xspace}
\begin{document}

\title{Real-Time Semantic Segmentation via Auto Depth, Downsampling Joint Decision and Feature Aggregation%\thanks{Grants or other notes
%about the article that should go on the front page should be
%placed here. General acknowledgments should be placed at the end of the article.}
}
% \subtitle{Do you have a subtitle?\\ If so, write it here}
%\titlerunning{Short form of title}        % if too long for running head

\author{Peng Sun   \and
        Jiaxiang Wu \and
        Songyuan Li \and
        Peiwen Lin \and
        Junzhou Huang \and
        Xi Li 
         %etc.
}

%\authorrunning{Short form of author list} % if too long for running head

\institute{Peng Sun \at
              Zhejiang University \\
              \email{sunpeng1996@zju.edu.cn}           %  \\
%             \emph{Present address:} of F. Author  %  if needed
           \and
           Jiaxiang Wu \at
              Tencent AI Lab 
%              \email{jonathanwu@tencent.com}
           \and
           Songyuan Li \at
              Zhejiang University 
           \and
           Peiwen Lin \at
              Sensetime Research 
           \and
           Junzhou Huang \at
              University of Texas at Arlington
%              \email{jzhuang@uta.edu}
           \and
           Xi Li \at
           Corresponding Author \\
              Zhejiang University \\
              \email{xilizju@zju.edu.cn}
}

\maketitle

\begin{abstract}

To satisfy the stringent requirements on computational resources in the field of real-time semantic segmentation, most approaches focus on the hand-crafted design of light-weight segmentation networks. Recently, Neural Architecture Search~(NAS) has been used to search for the optimal building blocks of networks automatically, but the network depth, downsampling strategy, and feature aggregation way are still set in advance by trial and error. In this paper, we propose a joint search framework, called AutoRTNet, to automate the design of these strategies. Specifically, we propose hyper-cells to jointly decide the network depth and downsampling strategy, and an aggregation cell to achieve automatic multi-scale feature aggregation. Experimental results show that AutoRTNet achieves 73.9\% mIoU on the Cityscapes test set and 110.0 FPS on an NVIDIA TitanXP GPU card with $768 \times 1536$ input images.

% \PACS{PACS code1 \and PACS code2 \and more}
% \subclass{MSC code1 \and MSC code2 \and more}
\end{abstract}

\section{Introduction}
\label{intro}
Semantic segmentation, a fundamental topic in computer vision, aims at assigning per-pixel semantic labels for images. Recent approaches~\cite{zhao2017pyramidpspnet,chen2017rethinkingdeeplabv3,chen2018encoderdeeplabv3plus,zhao2018psanet} based on fully convolutional networks~\cite{long2015fullyfcn} have achieved remarkable accuracy on public benchmarks \cite{brostow2008segmentationcamvid,cordts2016cityscapes,everingham2015pascal}. Such improvements, however, come at the cost of deeper and less efficient networks, which may not be applicable to many real-time systems, \eg autonomous driving and video surveillance.

\begin{figure}[t]
\centering
\includegraphics[width=8.0cm]{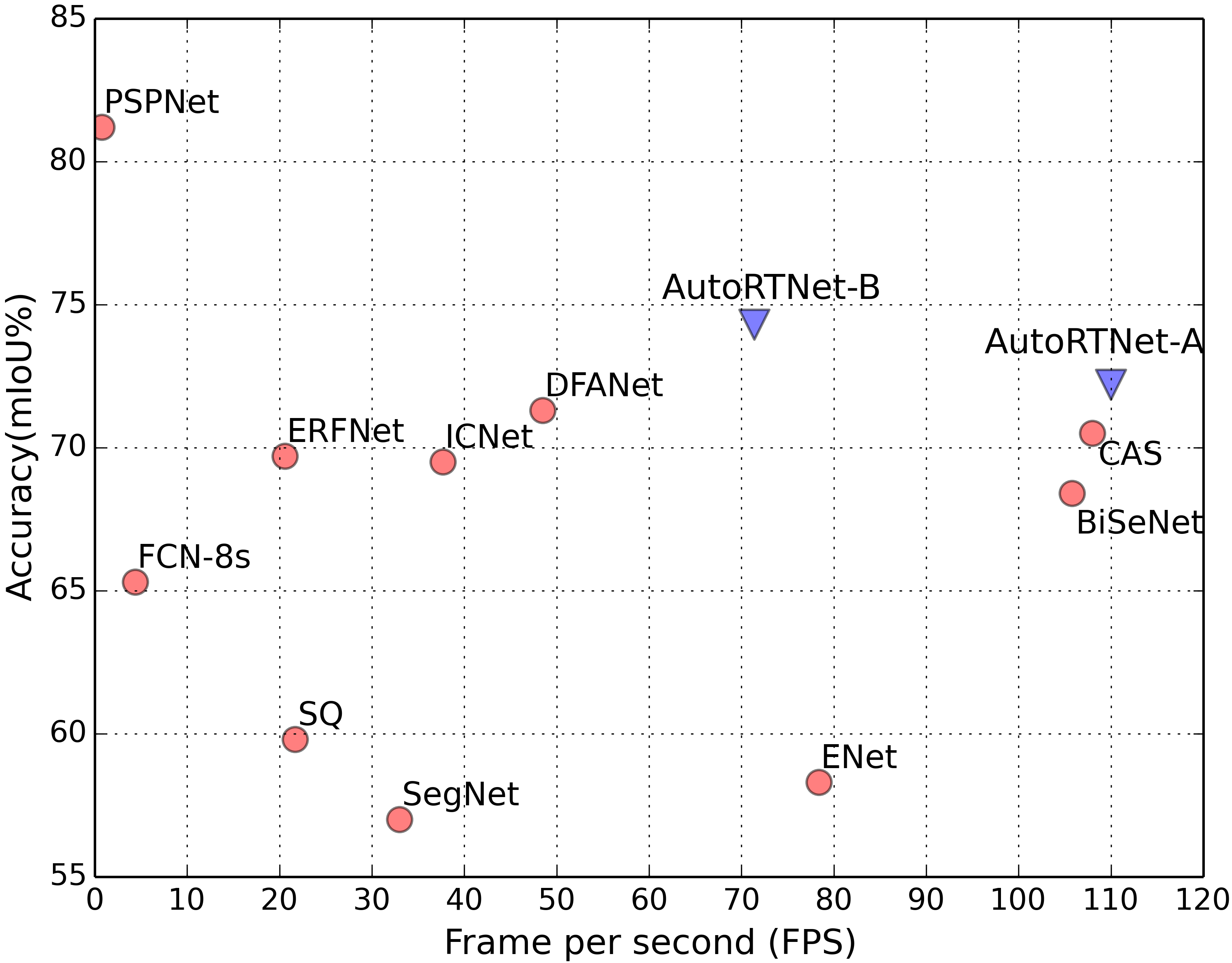}
\caption{The inference speed and accuracy for different networks on the Cityscapes test set. Compared with other methods, our AutoRTNet locates in the right-top since it features lower latency with comparable accuracy. \textbf{Best viewed in color.}}
\label{fig:plot}
\end{figure}

To perform fast semantic segmentation with satisfactory accuracy, the design philosophy of real-time segmentation network architectures mainly concentrates on three aspects: 1) building block design \cite{li2019dabnet,paszke2016enet}, which considers the block-level feature representation capacity, computational complexity, and receptive field size; 2) network depth and downsampling strategy~\cite{li2019dabnet,li2019dfanet}, which directly affect the accuracy and speed of networks, hence real-time networks favor shallow layers and fast downsampling; and 3) feature aggregation~\cite{yu2018bisenet,zhao2018icnet}, which fuses multi-scale features to compensate the loss of spatial details caused by fast downsampling.

The above hand-crafted networks make huge progress, while they require expertise in architecture design based on laborious trial and error. To relieve this burden, some researchers introduce neural architecture search (NAS) methods \cite{baker2016designing,zoph2016neural,DBLP:conf/iclr/LiuSY19,DBLP:conf/iclr/XieZLL19} into this field, and obtain excellent results~\cite{chen2018searchingdpc,liu2019autodeeplab,zhang2019customizable,nekrasov2019fastshenchunhua}. AutoDeepLab \cite{liu2019autodeeplab} and DPC \cite{chen2018searchingdpc} focus on high-quality segmentation instead of real-time applications. To meet the real-time demand, CAS \cite{zhang2019customizable} searches a customized architecture by introducing a latency loss function. Though its building block is searched, the network depth, downsampling strategy and feature aggregation way are still set by hand in advance and non-adjustable during the search process. While the aforementioned three aspects are highly correlated and indispensable for a remarkable real-time segmentation network, and these non-adjustable settings increase the difficulties and limitations to find an optimal real-time segmentation architecture (\ie the best trade-off between performance and speed). These motivate us to explore all of the aspects automatically during the search process.

In this paper, we propose a joint search framework to search for the optimal building blocks, network depth, downsampling strategy, and feature aggregation way simultaneously. Specifically, we propose hyper-cells to jointly decide the network depth and downsampling strategy via a cell-level pruning process in an adaptive manner, and an aggregation cell to fuse features from multiple spatial scales automatically. As for the hyper-cell, we introduce a novel learnable architecture parameter for it, and the network depth and downsampling strategy are fully determined concurrently according to the optimized hyper-cell architecture parameters.
As for the aggregation cell, we aggregate multi-level features in the network automatically to effectively fuse the low-level spatial details and high-level semantic context.

We denote the resulting network as \textbf{Auto} searched \textbf{R}eal-\textbf{T}ime semantic segmentation network or \textbf{AutoRTNet}. We evaluate AutoRTNet on both Cityscapes \cite{cordts2016cityscapes} and CamVid \cite{brostow2008segmentationcamvid} datasets. The experiments demonstrate the superiority of AutoRTNet, as shown in Figure \ref{fig:plot}, where our AutoRTNet achieves the best accuracy-efficiency trade-off. 
%Compared to other state-of-the-art methods, AutoRTNet achieves the new best trade-off between accuracy and speed on both datasets.

The main contributions can be summarized as follows:

\begin{itemize}
    		\item We propose a joint search framework for real-time semantic segmentation that automatically searches for the building blocks, network depth, downsampling strategy, and feature aggregation way simultaneously.
    		
    		\item We propose the hyper-cell to learn the network depth and downsampling strategy in an adaptive manner via the cell-level pruning process, and the aggregation cell to achieve automatic multi-scale feature aggregation.
  
       	    \item Notably, AutoRTNet has achieved 73.9\% mIoU on the Cityscapes test set and 110.0 FPS on an NVIDIA TitanXP GPU card with $768 \times 1536$ input images.
\end{itemize}

 \section{Related Work}

\begin{figure*}
\begin{center}
\includegraphics[width=7in]{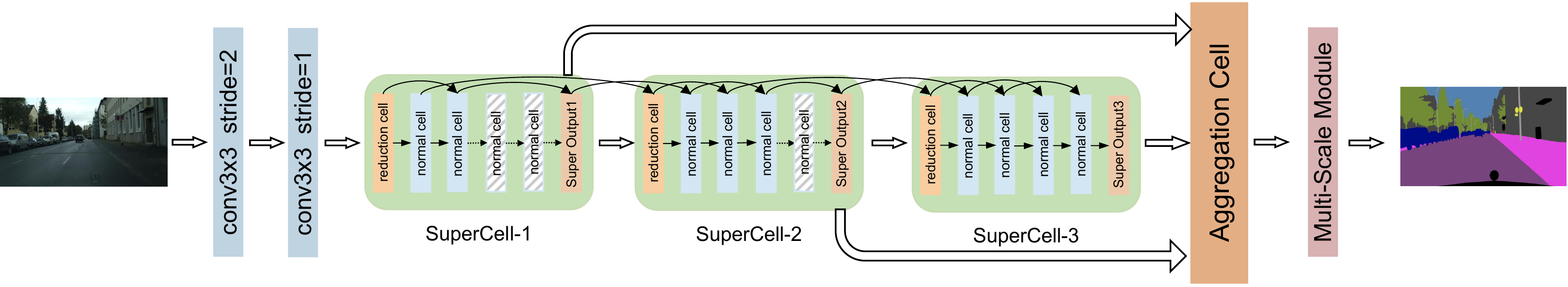}
\end{center}
 \caption{Illustration of our joint network architecture search framework. The network begins with two convolution layers and contains three hyper-cells which search for the optimal network depth and downsampling strategy via the cell-level pruning process, each hyper-cell contains a reduction cell and $n$ normal cells. The cells marked with the dotted white line are pruned after optimization. The aggregation cell is designed to perform automatic multi-scale feature aggregation effectively, and it seamlessly integrates the outputs of hyper-cells.}
\label{main}
\end{figure*} 

\label{sec:1}
\subsection{Semantic Segmentation}

\paragraph{High-Quality Segmentation} FCN \cite{long2015fullyfcn} is the pioneer work which has greatly promoted the development of semantic segmentation. Extensions to FCN follow many directions. Encoder-decoder structures \cite{badrinarayanan2017segnet,lin2017refinenet,noh2015learningdeconvolution} combine low-level and high-level features to improve accuracy of semantic segmentation. DRN \cite{yu2017dilated} and DeepLab \cite{chen2017rethinkingdeeplabv3,chen2018encoderdeeplabv3plus} use dilated convolution to effectively enlarge the receptive field. To capture multi-scale context information, DeepLabV3 \cite{chen2017rethinkingdeeplabv3} and PSPNet \cite{zhao2017pyramidpspnet} propose the pyramid modules. Recently, attention mechanism \cite{vaswani2017attention} has been used in segmentation field \cite{fu2019dualattention,zhang2019co,zhao2018psanet,li2018pyramid}. These outstanding works are designed for high-quality segmentation, which is inapplicable to real-time applications.

\paragraph{Real-Time Methods} Various algorithms have been proposed for real-time semantic segmentation. Some works \cite{wu2017realresize} reduce the computation overheads via restricting the input image size, channel-pruning algorithms \cite{paszke2016enet,badrinarayanan2017segnet} are introduced to boost the inference speed, and most real-time methods focus on designing the light-weight and effective network architectures. The design philosophy mainly can be summarized into the following three aspects. And in our work, we fully explore all three aspects simultaneously.

\noindent\textbf{\emph{Building block design}} The building block design \cite{paszke2016enet,romera2017erfnet,mehta2018espnet,li2019dabnet} requires researchers to give sufficient consideration to the computational complexity, feature representation capacity, and receptive field size, which is essential for real-time semantic segmentation. For example, ENet \cite{paszke2016enet} and DABNet \cite{li2019dabnet} propose light-weight blocks and stack them with different dilation rates to form a whole network. MobileNet and its variants \cite{howard2017mobilenets,sandler2018mobilenetv2} use blocks with depth-wise separable convolution in pursuit of light-weight models.

\noindent\textbf{\emph{Network depth and downsampling strategy}} Different from high-quality segmentation networks using pre-defined backbones (ResNet \cite{he2016deepresnet}, Xception \cite{chollet2017xception}, etc.) as encoders, the network depth and downsampling strategy (\ie how many blocks or layers in each stage) are determined mostly by hand for real-time segmentation networks (\eg DABNet \cite{li2019dabnet}, DFANet \cite{li2019dfanet}, ERFNet \cite{romera2017erfnet}), as they directly affect the accuracy and speed of the networks. For pursuing fast inference speed, real-time networks always enjoy shallow layers and perform fast downsampling with factor 16 or 32.

\noindent\textbf{\emph{Feature aggregation}} The fast downsampling of real-time networks easily results in the loss of spatial details. Thus, multi-scale feature aggregation \cite{yu2018bisenet,zhao2018icnet,li2019dfanet} has been proposed to remedy the loss of spatial details. ICNet \cite{zhao2018icnet} proposes a image cascade network with multi-scale inputs. BiSeNet \cite{yu2018bisenet} decouples the network into context and spatial paths to make a right balance between the accuracy and speed. DFANet \cite{li2019dfanet} aggregates multi-scale features from different layers to remedy the loss of spatial details.
\label{sec:2}
\subsection{Neural Architecture Search}

\paragraph{NAS Overview} Neural architecture search (NAS) focuses on automating the network architecture design process. Early NAS methods are time-consuming (\eg thousands of GPU days) and computationally expensive via reinforcement learning \cite{zoph2016neural,baker2016designing,zoph2018learningscalable,tan2019mnasnet} or evolutionary algorithms \cite{miikkulainen2019evolving,real2019regularized}. Recently, the emergence of differentiable NAS methods \cite{DBLP:conf/iclr/LiuSY19,DBLP:conf/iclr/XieZLL19,cai2018proxylessnas} has greatly relieved the time-consuming problem while achieving excellent performance. DARTs \cite{DBLP:conf/iclr/LiuSY19} is the pioneer work for gradient-based NAS, they propose an iterative optimization framework which is based on the continuous relaxation of the architecture representation. SNAS \cite{DBLP:conf/iclr/XieZLL19} constrains the architecture parameters to approximate one-hot, resolving the inconsistency in optimizing between the performance of derived child networks and converged parent networks. FBNet \cite{wu2019fbnet}, ProxylessNAS \cite{cai2018proxylessnas}, MnasNet \cite{tan2019mnasnet} propose multi-objective optimization with the consideration of real-world latency.

\paragraph{NAS For Segmentation} DPC \cite{chen2018searchingdpc} is the first work for dense image prediction using NAS methods and searches for a multi-scale representation module. The similar work to us is AutoDeepLab \cite{liu2019autodeeplab}, they propose a hierarchical search space and search for the downsampling path. Although they also search for the downsampling strategy, the mechanism is extremely different from ours. They design the network level continuous relaxation to learn the downsampling path, while we search for the downsampling strategy via the cell-level pruning progress. Moreover, they cannot search for the adaptive network depth and feature aggregation way and they focus on high-quality segmentation instead of real-time applications. For real-time requirements, CAS \cite{zhang2019customizable} searches for an architecture with customized resource constraints and achieves excellent real-time performance. However, our approach also searches for the adaptive network depth, downsampling strategy and feature aggregation way, which is extremely different from CAS \cite{zhang2019customizable}.

\paragraph{NAS For Object Detection} The combination of multi-scale features is also essential for object detection \cite{lin2016fpn,Liu2016SSDSS}. In the field of NAS, NAS-FPN \cite{Ghiasi_2019_CVPR_NAS-FPN} and Auto-FPN \cite{Xu_2019_ICCV_Auto-FPN} also search for an architecture that merges features of varying dimensions and are successful at searching for the appropriate combination method. Unlike us, NAS-FPN \cite{Ghiasi_2019_CVPR_NAS-FPN} proposes merging cells and uses RNN controller to select candidate feature layers and a binary operation in each merging cell. Their search space only consists of two binary operations, \ie sum and global pooling for their simplicity. Auto-FPN \cite{Xu_2019_ICCV_Auto-FPN} searches for an efficient feature fusion module, which search space is specially designed for detection and flexible enough to cover many popular designs of detectors. Thus the search space design, motivation and implementation of above both methods are different from ours.

 \section{Methods}

The joint search framework is shown in Figure \ref{main}.  We propose the hyper-cell to search for the optimal network depth and downsampling strategy as they directly affect the accuracy and speed of networks. For remedying the loss of spatial details caused by fast downsampling, a novel aggregation cell is proposed for automatic multi-scale feature aggregation. The framework contains two pre-defined convolution layers, three hyper-cells and an aggregation cell. The multi-scale module \cite{zhang2019customizable} is subsequently used to extract the global and local context for final prediction. For real-time demands, we take the real-world latency into consideration during the search process. We begin this section with the differentiable architecture search. Afterwards, we elaborate the proposed hyper-cell and aggregation cell in detail.

\subsection{Differentiable Architecture Search\label{method3-1}}

\paragraph{Intra-cell Search space}

The hyper-cell is the building block of the network, and the cell is the basic component unit of the hyper-cell, as shown in Figure \ref{main}. There are two types of cells, normal cells and reduction cells \cite{DBLP:conf/iclr/LiuSY19,DBLP:conf/iclr/XieZLL19}. The reduction cells reduce the feature map size by a factor of 2 for downsampling, and the factor is 1 in normal cells.

A cell is a directed acyclic graph (DAG) consisting of an ordered sequence of $N$ nodes, denoted by $\mathcal{N}$ = $\{x^{(1)},...,x^{(N)}\}$. Each node $x^{(i)}$ is a latent representation (\ie feature map), and each directed edge $\left(i, j \right)$ is associated with some candidate operations (\eg conv, pooling) in operation set $\mathcal{O}^{(i, j)}$, representing all possible transformations from $x^{(i)}$ to $x^{(j)}$. Each cell has two inputs (the outputs of the previous two cells) and one output (the concatenation of all the intermediate nodes in the cell). The structure of cell is shown on the right in Figure \ref{supercell}. Each intermediate node $x^{(j)}$ is computed based on all of its predecessors:

\begin{equation}
% \begin{gather*}
x^{(j)} = \sum\nolimits_{i < j} \widetilde{o}^{(i, j)} \big( x^{(i)} \big)
% \end{gather*}
\end{equation}

where $\widetilde{o}^{(i, j)}$ $\in$ $\mathcal{O}^{(i, j)}$ is the optimal operation at edge $(i,j)$. 

In order to determine the optimal operation $\widetilde{o}^{(i, j)}$ at edge $(i,j)$, we represent the intra-cell search space with a set of one-hot random variables from a fully factorizable joint distribution $p(M)$ \cite{DBLP:conf/iclr/XieZLL19}. Specifically, each edge ($i$, $j$) is associated with a one-hot random variable $M^{(i,j)}$. We use $M^{(i,j)}$ as a mask to multiply all the candidate operations $\mathcal{O}^{(i, j)}$ at edge $(i,j)$, and the intermediate node $x^{(j)}$ is given by: 

\begin{equation}
x^{(j)} =  \sum\nolimits_{i < j} \sum\nolimits_{o \in \mathcal{O}} m^{(i,j)}_{o} \cdot o^{(i, j)} \big( x^{(i)} \big)
\end{equation}

% \begin{equation}
% x^{(j)} =  \sum\nolimits_{i < j} \sum\nolimits_{o \in \mathcal{O}} m^{(i,j)}_{o} \cdot o^{(i, j)} \big( x^{(i)} \big)
% \end{equation}

where $m^{(i,j)}_{o}$ $\in$ $M^{(i,j)}$ and $m^{(i,j)}_{o}$ is a random variable in $\{0, 1\}$, it is evaluated to 1 if operation $o^{(i, j)}$ is selected.

To make $p(M)$ differentiable, we use Gumbel Softmax technique \cite{jang2016categorical,maddison2016concrete} to relax the discrete sampling distribution to be continuous and differentiable:
\begin{equation}
\label{equ:softmax}
M^{(i,j)} = f_{\alpha^{(i,j)}}(G^{(i,j)}) = \text{softmax} ((\log \alpha^{(i,j)} + G^{(i,j)}) / \lambda)
\end{equation}
where $M^{(i,j)}$ is the softened one-hot random variable for operation selection at edge ($i$, $j$), $\alpha^{(i,j)}$ is the intra-cell architecture parameter at edge ($i$, $j$), $G^{(i,j)}$ = $-log(-log(U^{(i,j)}))$ is a vector of Gumbel random variables, $U^{(i,j)}$ is a uniform random variable in the range $(0, 1)$. $\lambda$ is the temperature of softmax, and as $\lambda$ approaches 0, $M^{(i,j)}$ approximately becomes one-hot. The technique of using Gumbel Softmax makes the entire intra-cell search differentiable \cite{wu2018mixed,wu2019fbnet,DBLP:conf/iclr/XieZLL19} to both network parameter $w$ and architecture parameter $\alpha$.

For the candidate operation set $\mathcal{O}$, we collect the operations as follows:

% \vspace{-0.1in}
\begin{itemize}[itemsep=0pt, parsep=0pt]
\footnotesize
\item zero operation
\item skip connection
\item 3 $\times$ 3 max pooling
\item 3 $\times$ 3 conv
\item 3 $\times$ 3 conv, repeat 2
\item 3 $\times$ 3 separable conv
\item 3 $\times$ 3 separable conv, repeat 2 
\item 3 $\times$ 3 dilated separable conv, dilation=2 
\item 3 $\times$ 3 dilated separable conv, dilation=4 
\item 3 $\times$ 3 dilated separable conv, dilation=2, repeat 2
\end{itemize}

\paragraph{Intra-cell Latency Cost}

For the operation selection of cells towards real-time network, we take real-world latency into consideration. Specifically, we build a GPU-latency lookup table \cite{cai2018proxylessnas,tan2019mnasnet,wu2019fbnet,zhang2019customizable} that records the inference time cost of each candidate operation. The latency of each operation is measured in micro-second on a TitanXP GPU. During the search process, we associate a cost $lat_{o}^{(i,j)}$ with each candidate operation $o^{(i,j)}$ at edge $(i, j)$, thus the latency cost of cell $p$ is formulated as:
\begin{equation}\label{equ:latency}
lat_p \! = \!  \sum\nolimits_{(i,j)} \sum\nolimits_{o \in \mathcal{O}} m^{(i,j)}_{o} \cdot lat_{o}^{(i,j)}
\end{equation}
where $m^{(i,j)}_{o} \in M^{(i,j)}$ and $M^{(i,j)}$ denotes the softened one-hot random variable at edge $(i,j)$. By using the pre-built lookup table and above sampling process, the latency loss is also differentiable with respect to $m^{(i,j)}_{o}$.

\subsection{Adaptive Network Depth and Downsampling}

\paragraph{Hyper-Cell Search Space}

\begin{figure}[t]
\centering
\includegraphics[width=8cm]{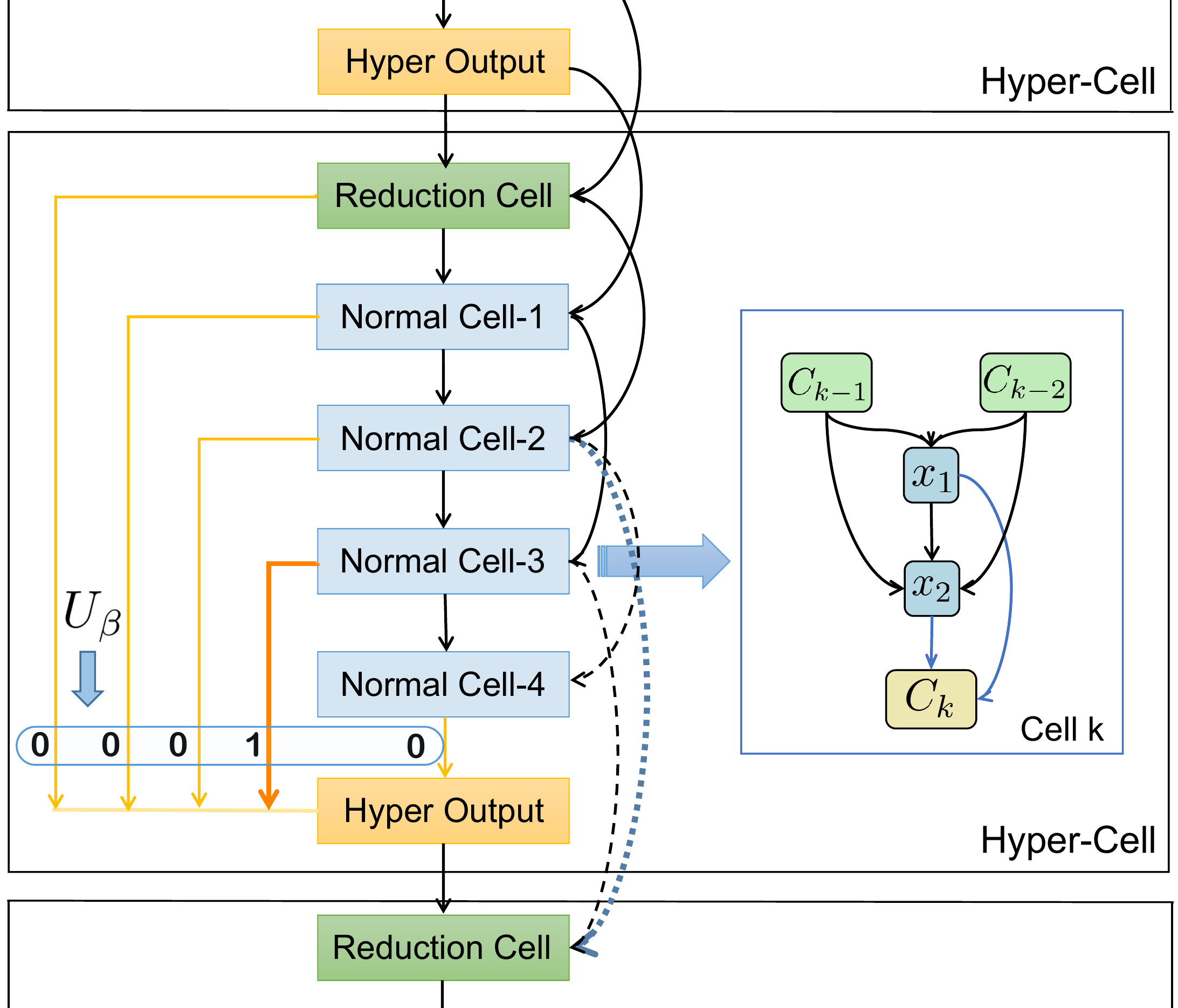}
\caption{Illustration of our hyper-cell. The hyper-cell consists of a reduction cell and $n$ normal cells and $n + 1$ edges with architecture parameter which encodes the depth of the hyper-cell. The structure of cell is shown on the right in this figure.}
\label{supercell}
\end{figure}

The network depth and downsampling strategy directly affect the accuracy and speed of networks for real-time semantic segmentation. To adjust them in an adaptive manner, we formulate the two design-making processes as a single cell-level pruning process. Specifically, we propose a hyper-cell, as shown in Figure \ref{supercell}, which consists of a reduction cell and $n$ normal cells. We introduce $n + 1$ edges to connect each cell with the hyper-cell's output, and associate them with the learnable architecture parameter $\beta$. The intra-cell architecture parameters $\alpha$ of $n$ normal cells are shared in the same hyper-cell.

We determine the depth of each hyper-cell by limiting that only one edge can be activated for each hyper-cell, and all cells behind this activated edge can be pruned safely. Each specific edge in hyper-cell $s$ is associated with a one-hot random variable $U^s$ = ($u_1^s$, $u_2^s$, \dots, $u_{n+1}^s$) from a fully factorizable joint distribution $P(U)$. The $U^s$ works as a mask during the training process, and the output of the hyper-cell $s$ is designed as:
\begin{equation}
HyperOut^{(s)} = \sum\nolimits_{p=1}^{n+1} {u_{p}} ^ {s} \cdot ({C_{p}} ^ {s})
\end{equation}
where $C_{p}^ {s} $ is the output of $p$-th cell in hyper-cell $s$, $u_{p}^{s}$ represents the random variable in $\{0, 1\}$ of $p$-th edge of hyper-cell $s$. Similarly, we adopt the Gumbel Softmax based sampling process to make the training process differentiable.
\begin{equation}\label{equ:softmaxtwo}
U^{s} \! = f_{\beta^{s}}(G^{s}) = \text{softmax} ((\log \beta^{s} + G^{s}) / \lambda)
\end{equation}
where $U^{s}$ is the softened one-hot random variable for edge selection of hyper-cell $s$, $\beta^{s}$ is the architecture parameter of hyper-cell $s$. $G^{s}$ and $\lambda$ are similar to the ones in equation (\ref{equ:softmax}). The hyper-cell architecture parameter $\beta$ we introduced can be effectively optimized together with the network parameter $w$, intra-cell architecture parameter $\alpha$ in the same round of back-propagation. After stacking hyper-cells to form a whole network, the network depth and downsampling strategy can be fully explored concurrently according to the architecture parameter $\beta$.

To better explain the cell-level pruning process, we give an example as follows. In the initial phase, let's say we have five cells (one reduction cell and four normal cells) and each cell in hyper-cell keeps its original inputs and outputs. As shown in Figure \ref{supercell}, if the fourth edge is activated currently (\ie $U$ is $\{0, 0, 0, 1, 0\}$), the Normal Cell-4 will be pruned in this iteration, and the output of this hyper-cell is the output of Normal Cell-3. At the same time, the reduction cell in next hyper-cell $s+1$ takes the outputs of hyper-cell $s$ and Normal Cell-2 in hyper-cell $s$ as its inputs, to stick to the ``two-input" principle of the cell. The learning and adjusting like this go through the entire searching phase.

By introducing the architecture parameter $\beta$ in the proposed hyper-cell, we can dynamically adjust and search for the network depth as well as the downsampling strategy for real-time semantic segmentation.

\paragraph{Network Latency Cost} We define the set of cells in all hyper-cells in the initial phase as $P$, after optimization, the number of the set is reduced and the new set is marked as $\bar{P}$. For the current architecture $(\alpha,\beta)$ containing several hyper-cells, the total latency excludes the pruned cells and can be calculated as:

\begin{equation}\label{equ:latency2}
Latency(\alpha,\beta)  =  \sum\nolimits_{p \in \bar{P} } lat_{p}
\end{equation}
where $\bar{P}$ is the set of remaining cells in all hyper-cells of architecture $(\alpha,\beta)$. The $lat_{p}$ is the latency of cell $p$. Thus the total loss function is formulated as:

\begin{equation}\label{equ:latency2}
L((\alpha,\beta), w)  = CE((\alpha,\beta), w) + \gamma ~log(Latency((\alpha,\beta)))
\end{equation}
where $CE((\alpha,\beta), w)$ is the cross-entropy loss of architecture $(\alpha,\beta)$ with network weight $w$ and $\gamma$ controls the magnitude of latency term (\ie balance the trade-off between accuracy and speed).

\subsection{Network-Level Auto Feature Aggregation}

\begin{figure}[t]
\centering
\includegraphics[width=8.5cm]{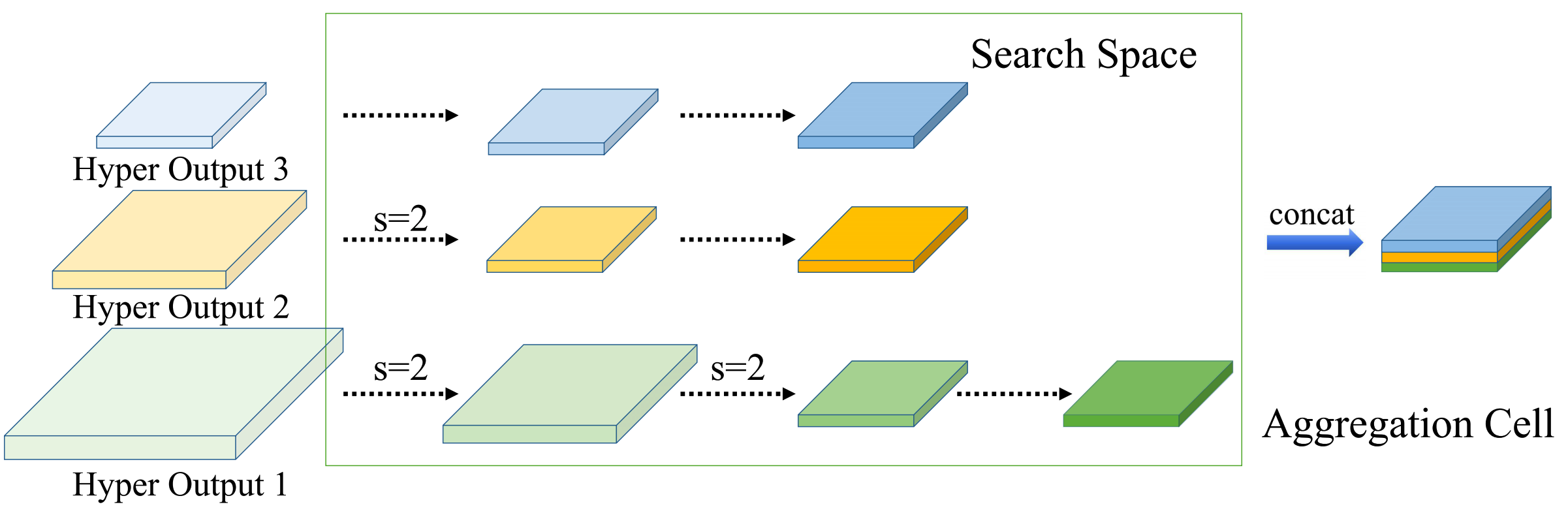}
\caption{Overview of the aggregation cell for automatic multi-scale feature aggregation. The aggregation cell contains $E$ edges (dotted arrows), each edge is equiped with some candidate operations. The ``s=2'' means stride = 2. \textbf{Best viewed in color.}}
\label{aggregation}
\end{figure}

For remedying the loss of spatial details in real-time segmentation networks due to fast downsampling, we propose the aggregation cell to automatically aggregate features by optimal operations from different levels in the network. The aggregation cell seamlessly integrates the outputs of above hyper-cells, and the outputs of the early hyper-cells compensate for the loss of spatial details.

The structure of the proposed aggregation cell is shown in Figure \ref{aggregation}. The aggregation cell takes three hyper-cells' outputs with different resolutions as its inputs, thus the aggregation cell is designed to combine multi-scale features (\ie low-level spatial details and high-level semantic context). The aggregation cell is designed as a directed acyclic graph consisting of $M$ nodes and $E$ edges, each node is a latent representation (\ie feature map) and each directed edge is associated with some candidate operations. As shown in Figure \ref{aggregation}, each edge's stride is set to 1, unless explicitly specified by ``s=2'' (stride 2), which acts as the downsampling connection. The output of the aggregation cell is designed as the concatenation of the final feature maps from three hyper-cells. We use the same sampling and optimization process as intra-cell search in Section \ref{method3-1} to optimize the aggregation cell's architecture parameter.

Given the candidate operation set, the aggregation cell also efficiently enlarges the receptive field of the network. Moreover, the aggregation cell is designed for effectively improving the segmentation accuracy, thus we introduce it without the latency constraint. For the operation set of the aggregation cell, we collect following 5 kinds of operations:

% \vspace{-0.06in}
\begin{itemize}[itemsep=0pt, parsep=0pt]
\footnotesize
\item 1$\times$1 conv, repeat 2
\item 3$\times$3 conv, repeat 2
\item 3$\times$3 dilated separable conv, dilation=2, repeat 2
\item 3$\times$3 dilated separable conv, dilation=4, repeat 2
\item 3$\times$3 dilated separable conv, dilation=8, repeat 2
\end{itemize}
\label{Section 3.3}
% \vspace{-0.2in}

%------------------------------------------------------------------------
\section{Experiments}

To verify the effectiveness and superiority of our joint search framework, we compare our AutoRTNet with other state-of-the-art methods on two challenging benchmark datasets: Cityscapes \cite{cordts2016cityscapes} and CamVid \cite{brostow2008segmentationcamvid}. Moreover, we conduct a series of ablation studies to verify the effectiveness of the proposed hyper-cell and aggregation cell. Finally, we provide an in-depth analysis about the detailed architecture of AutoRTNet.

\subsection{Implementation Details}

\paragraph{Searching} For the searching process, the whole network contains three hyper-cells and the initial cell numbers in these hyper-cells are $\{5, 10, 10\}$, respectively. The intermediate node number of the cell is set to 2. The initial channel number is 8, and the channels are $\times$3 when downsampling in reduction cells. The search process, which is conducted on the Cityscapes dataset, runs 150 epochs with mini-batch size 16, which takes approximately 16 hours with 16 TitanXP GPU cards. Similar to FBNet \cite{wu2019fbnet}, we postpone the training of the hyper-cell architecture parameter $\beta$ by 50 epochs to warm-up weight $w$ and intra-cell architecture parameter $\alpha$. The $\alpha$ and $\beta$ are optimized by Adam, with initial learning rate 0.001, momentum (0.5, 0.999) and weight decay 1e-4. The $w$ is optimized using SGD with momentum 0.9, weight decay 1e-3, and cosine learning scheduler that decays learning rate from 0.025 to 0.001. For Gumbel Softmax, we empirically set the initial temperature $\lambda$ in equation (\ref{equ:softmax}) and (\ref{equ:softmaxtwo}) as 3.0, and gradually decrease to the minimum value of 0.03. We set the node number $M$ and edge number $E$ as 7 in the aggregation cell. 

\paragraph{Retraining} When the search process is over, the searched network is firstly pretrained on the ImageNet dataset from scratch. We then finetune the network on the specific segmentation dataset (\ie Cityscapes or CamVid) for 200 epochs with mini-batch size 16. The base learning rate is 0.01 and the ‘poly’ learning rate policy is adopted with power 0.9, together with momentum is 0.9 and weight decay is 0.0005. Following \cite{Wu2016High,yu2018bisenet}, we compute the loss function with the online bootstrapping strategy. Data augmentation contains random horizontal flip, random resizing with scale ranges in [0.5, 2.0], and random cropping into fix size for training.

\subsection{Benchmarks and Evaluation Metrics}

Cityscapes \cite{cordts2016cityscapes}, a public street scene dataset, contains high quality pixel-level annotations of 5000 images with size 1024 $\times$ 2048 and 19,998 images with coarse annotations. 19 semantic classes are used for training and evaluation. CamVid \cite{brostow2008segmentationcamvid} is another public dataset, and it contains 701 images in total. We follow the training/testing set split in \cite{zhang2019customizable,brostow2008segmentationcamvid}, with 468 training and 233 testing labeled images. These images are densely labeled with 11 semantic class labels. We use three evaluation metrics, including mean of class-wise intersection over uniou (mIoU), network forward time (Latency), and Frames Per Second (FPS).

\subsection{Real-time Semantic Segmentation Results}

In this section, we compare the AutoRTNet with other real-time segmentation methods. We run all experiments based on Pytorch 0.4 \cite{paszke2017automaticpytorch} and measure the latency on an NVIDIA TitanXP GPU card under CUDA 9.0. For fair comparison, we directly quote the reported remeasured or estimated speed results on TitanXP of other algorithms mentioned in \cite{zhang2019customizable,orsic2019defenseSwiftNet}. For the AutoRTNet, we report the average inference time through 500 times. In this process, we don't employ any test augmentation. 

%%% cityscapes
\begin{table}[h]
\caption{Accuracy and speed comparison of our method against other state-of-the-art methods on Cityscapes test set. Methods trained using both fine and coarse data are marked with $*$. The mark ${\dag}$ represents the speed is remeasured by us on Titan XP.}
\begin{center}
\setlength{\tabcolsep}{1mm}{
\begin{tabular}{|l|c|c|c|c|}
\hline
Method  & Input Size & mIoU (\%) & Latency(ms) & FPS \\
\hline\hline
FCN-8S \cite{long2015fullyfcn} & 512 $\times$ 1024  & 65.3 & 227.23 & 4.4    \\
PSPNet \cite{zhao2017pyramidpspnet} & 713 $\times$ 713 & 81.2 & 1288.0 & 0.78   \\
DeepLabV3$^*$ \cite{chen2017rethinkingdeeplabv3} & 769 $\times$ 769   & 81.3 & 769.23 & 1.3    \\
AutoDeepLab$^*$ $^{\dag}$\cite{liu2019autodeeplab} & 769 $\times$ 769 & 81.2 & 303.0 & 3.3    \\
\hline
SegNet \cite{badrinarayanan2017segnet} & 360 $\times$ 640 & 57.0 & 30.3   & 33     \\
ENet \cite{paszke2016enet} & 360 $\times$ 640   & 58.3 & 12.7   & 78.4   \\
SQ \cite{treml2016speedingSQ}  & 1024 $\times$ 2048 & 59.8 & 46.0   & 21.7   \\
ERFNet \cite{romera2017erfnet} & 512 $\times$ 1024 &  69.7 & 48.5  & 20.6 \\
ICNet \cite{zhao2018icnet}  & 1024 $\times$ 2048 & 69.5 & 26.5   & 37.7   \\
DF1-Seg \cite{li2019partialdfnet} & 768 $\times$ 1536 & 73.0 & 29.1 & 34.4   \\
SwiftNet \cite{orsic2019defenseSwiftNet}& 1024 $\times$ 2048  &  75.1 & 26.2 & 38.1 \\
ESPNet \cite{mehta2018espnet} & 512 $\times$ 1024 & 60.3 & 8.2 & 121.7   \\
DFANet \cite{li2019dfanet} & 1024 $\times$ 1024 & 71.3 & 10.0  & 100.0   \\
DFANet $^{\dag}$ \cite{li2019dfanet} & 1024 $\times$ 1024 & 71.3 & 20.6 $^{\dag}$ & 48.5 $^{\dag}$  \\
BiSeNet \cite{yu2018bisenet} & 768 $\times$ 1536  & 68.4 & 9.52   & 105.8  \\
CAS \cite{zhang2019customizable} & 768 $\times$ 1536  & 70.5 & 9.25   & 108.0 \\
CAS$^*$ \cite{zhang2019customizable} & 768 $\times$ 1536  & 72.3 & 9.25   & 108.0  \\
\hline
\textbf{AutoRTNet-A} & 768 $\times$ 1536 & 72.2 &  9.09 & 110.0  \\
\textbf{AutoRTNet-A$^*$} & 768 $\times$ 1536 & 73.9 &  9.09 & 110.0  \\
\textbf{AutoRTNet-B}  & 768 $\times$ 1536 & 74.3 &  14.0 & 71.4 \\
\textbf{AutoRTNet-B$^*$}  & 768 $\times$ 1536 & 75.8 & 14.0 & 71.4  \\
\hline
\end{tabular}}
\end{center}
\label{Cityscapes}
\end{table}

\paragraph{Results on Cityscapes.} AutoRTNet-A and AutoRTNet-B are searched with latency term weight $\gamma$ 0.01 and 0.001, respectively. We evaluate them on the Cityscapes test set. The validation set is added for training before submitting to online Cityscapes server. Following \cite{zhang2019customizable,yu2018bisenet}, we scale the resolution of the images from 1024 $\times$ 2048 to 768 $\times$ 1536 as inputs to measure the speed and accuracy. As shown in Table \ref{Cityscapes}, our AutoRTNet achieves the best trade-off between accuracy and speed. AutoRTNet-A yields 72.2\% mIoU while maintaining 110.0 FPS on the Cityscapes test set with only fine data and without any test augmentation. When the coarse data is added to the training set, the mIoU achieves 73.9\%, which is the state-of-the-art trade-off for real-time semantic segmentation. Compared with BiseNet \cite{yu2018bisenet} and CAS \cite{zhang2019customizable} which have comparable speed to us, AutoRTNet-A surpasses them by 3.8\% and 1.7\% in mIoU on the Cityscapes test set, respectively. Compared with other real-time segmentation methods (\eg ENet \cite{paszke2016enet}, ICNet \cite{zhao2018icnet}), our AutoRTNet-A surpasses them in both speed and accuracy by a large margin. Moreover, our AutoRTNet-B achieves 74.3\% and 75.8\% mIoU (+coarse data) on the Cityscapes test set with 71.4 FPS, which is also the state-of-the-art real-time performance. 

\paragraph{Results on CamVid.} To validate the transferability of the networks searched by our framework, we directly transfer AutoRTNet-A and AutoRTNet-B, which are obtained on Cityscapes, to the CamVid dataset, as reported in Table \ref{CamVid}. With 720 $\times$ 960 input images, AutoRTNet-A achieves 73.5\% mIoU on CamVid test set with 140.0 FPS, which is the state-of-the-art trade-off between accuracy and speed. AutoRTNet-B achieves 74.2\% mIoU with 82.5 FPS. We also conduct the architecture search on CamVid ($\gamma = 0.1$) and name the resulting network AutoRTNet-C. Notably, AutoRTNet-C achieves amazing 250.0 FPS while maintaining 68.6\% mIoU on the CamVid test set, which surpasses ICNet \cite{zhao2018icnet} (67.1\% mIoU with 34.5 FPS) and DFANet \cite{li2019dfanet} (64.7\% mIoU with 120 FPS) significantly.

\begin{table}[h]
\begin{center}
\setlength{\tabcolsep}{1mm}{
\begin{tabular}{|l|c|c|c|c|}
\hline
Method  & ~mIoU (\%) ~ & Latency(ms) & ~FPS~ & Parameters (M) \\
\hline\hline
SegNet \cite{badrinarayanan2017segnet} & 55.6 & 34.01 & 29.4  &  29.5 \\
ENet \cite{paszke2016enet} & 51.3 & 16.33 & 61.2  & {0.36} \\ 
ICNet \cite{zhao2018icnet} & 67.1 & 28.98 & 34.5   & 26.5 \\
BiSeNet \cite{yu2018bisenet} & 65.6 &  -  & -  & 5.8 \\
DFANet \cite{li2019dfanet} & 64.7 & 8.33  & 120 & 7.8  \\
CAS \cite{zhang2019customizable}& 71.2 & 5.92  & 169 & - \\
\hline
\textbf{AutoRTNet-A} & \textbf{73.5} & 7.14  & 140.0 & 2.5 \\
\textbf{AutoRTNet-B} & \textbf{74.2} & 12.1 & 82.5 & 3.9 \\
\hline
\textbf{AutoRTNet-C} & \textbf{68.6} & 4.0 & 250.0 & 1.4 \\
\hline
\end{tabular}}
\end{center}
\caption{Results on the CamVid test set with resolution 720 $\times$ 960.}
\label{CamVid}
\end{table}

\paragraph{Parameter Results} For many real-time applications on computationally limited mobile platforms, which have restrictive memory constraints, thus model size (number of parameters) is also an important consideration. Table \ref{CamVid} shows the results of our AutoRTNet and other methods on CamVid test set. With only 2.5 million parameters, our AutoRTNet-A achieves impressive accuracy (\ie 73.5\% mIoU) on the CamVid test set, which significantly outperforms existing real-time segmentation networks. The model sizes of AutoRTNet B and C are 3.9M and 1.4M, respectively.

\subsection{Ablation Study}

The contribution of each component is investigated in following ablation studies on Cityscapes validation set. The latency term weight $\gamma$ in equation (\ref{equ:latency2}) is set to 0.01 and all networks are firstly pretrained on ImageNet in following experiments for fair comparison, if not specially noted.

\subsubsection{Comparison with Random Search}

%\begin{table}[h]
%\caption{The optimization results of hyper-cells with different initial states and different random seeds.}
%\begin{center}
%\setlength{\tabcolsep}{1mm}{
%begin{tabular}{|c|c|c|c|}
%\hline
%Method  & ~mIoU (\%) ~ & Latency (ms) & ~FPS~ \\
%\hline
%AutoRTNet   & 72.9 & 9.09 & 110.0  \\
%\hline
%random search  & 66.7 \pm  2.5 & 11.4 \sim 16.2  &  87.5  \sim 61.2  \\
%\hline
%\end{tabular}}
%\end{center}
%\label{randomsearchcom}
%\end{table}

\begin{table}[h]
\caption{Comparison with random search on the Cityscapes validation set.}
\begin{center}
\setlength{\tabcolsep}{1mm}{
\begin{tabular}{|c|c|c|c|}
\hline
Method  & ~mIoU (\%) ~ & Latency (ms) & ~FPS~ \\
\hline
AutoRTNet   & 72.9 & 9.09 & 110.0  \\
\hline
random search  & 66.7 $\pm$ 2.5 & 11.4 $\sim$ 16.2  &  87.5  $\sim$ 61.2  \\
\hline
\end{tabular}}
\end{center}
\label{randomsearchcom}
\end{table}

As discussed in \cite{randomsearch,Sciuto2019Evaluatingrandom}, NAS is a specialized hyper-parameter optimization problem, while random search is a competitive baseline for the problem. We apply random search to semantic segmentation by randomly sampling ten architectures from our previously-defined search space. The whole search space contains intra-cell operation selection and hyper-cell depth decision, which is significantly challenging for random search to find a satisfactory network. As shown in Table \ref{randomsearchcom}, random search achieves average mIoU 66.7\% $\pm$ 2.5\% on Cityscapes validation set with ImageNet pretrained, which is substantially lower than our AutoRTNet. The results also demonstrate the effectiveness of our search algorithm.

\subsubsection{Hyper-Cell}

\paragraph{Robustness} Firstly, to verify the robustness of hyper-cell, we set different initial numbers of cells and different random seeds in the initialization phase. The network contains three hyper-cells and the initial cell numbers in hyper-cells are set to \{a, b, c\}, after optimization, the numbers of cells remaining in each hyper-cell are \{$\overline{a}$, $\overline{b}$, $\overline{c}$\}. As shown in Table \ref{supercella}, the experiments demonstrate that the hyper-cells are insensitive to both initial numbers of cells and random seeds, which verify the robustness of the hyper-cell.

\begin{table}[h]
\caption{The optimization results of hyper-cells with different initial states and different random seeds.}
\label{supercella}
\begin{center}
\scalebox{0.85}{
\setlength{\tabcolsep}{1mm}{
\begin{tabular}{|c|c|c|c|c|}
\hline
Random seed & Initial phase & After optimization & mIoU & Frames Per  \\
setting & \{a, b, c\} & \{$\overline{a}$, $\overline{b}$, $\overline{c}$\} & (\%) & Second (FPS) \\
\hline\hline
2 & \{5, 10, 10\} & \{2, 4, 6\} & 72.9 & 110.0\\
2 & \{5, 10, 15\} & \{2, 4, 6\} & 73.0 & 106.5\\
2 & \{5, 15, 15\} & \{1, 4, 6\} & 72.5 & 102.8 \\
2 & \{10, 10, 10\}& \{2, 4, 6\} & 72.8 & 112.3 \\
\hline
1 & \{5, 10, 10\} & \{2, 4, 6\} & 73.0 & 113.6 \\
3 & \{5, 10, 10\} & \{1, 4, 7\} & 72.8 & 107.9 \\
\hline
\end{tabular}}
}
\end{center}
\end{table}

\paragraph{Downsampling strategy} To demonstrate the superiority of the downsampling strategy searched by hyper-cells, we compare the random downsampling position settings with the searched one. The total cell number is 12 ($\overline{a}$+$\overline{b}$+$\overline{c}$) searched by our framework, we fix the searched cell structures and only random change the downsampling positions (x, y, z) for fair comparison. The (x, y, z) represents the index positions of reduction cells in total 12 cells. After pretaining and retraining, the results in Table \ref{downsampling} demonstrate the superiority of the searched downsampling strategy through hyper-cells. Compared with the random ones, our hyper-cell achieves the best trade-off between accuracy and speed.

\begin{table}[h]
\caption{Comparison to random downsampling strategy.}
\label{downsampling}
\begin{center}
\scalebox{0.92}{
\setlength{\tabcolsep}{1mm}{
\begin{tabular}{|c|c|c|c|}
\hline
Downsampling Positions & mIoU(\%) & FPS & Downsampling Design Rule \\
\hline\hline
(1, 3, 7)   & 72.9 & 106 & Hyper Cell \\
(1, 5, 10) & 71.6 & 90.2 & Random \\
(3, 5, 8) & 72.5 & 75.7 & Random \\
(1, 2, 4)  & 68.4 & 125 & Random \\
\hline
\end{tabular}}
}
\end{center}
\end{table}
%\vspace{-3in}
% \vspace{-0.5in}

\subsubsection{Aggregation Cell}

To demonstrate the effectiveness of the proposed aggregation cell, we conduct a series of experiments with different strategies: a) without multi-scale feature aggregation; b) with random designed aggregation cell using random selected operations from the aggregation cell's search space; c) with searched aggregation cell (\ie our AutoRTNet-A). Among them, the result of the random aggregation cell is the average result over ten repeated random experiments and the results are shown in Table \ref{aggregationcell}. Overall, the searched aggregation cell successfully boosts up the mIoU from 69.9\% to 72.9\% on Cityscapes val set. Particularly, the searched aggregation cell surpasses the random one 1.5\% performance gains.

\begin{table}[h]
\caption{Ablation study for effectiveness of the aggregation cell.}
\begin{center}
% \scalebox{0.9}{
\setlength{\tabcolsep}{1mm}{
\begin{tabular}{|l|c|}
\hline
Methods  & mIoU (\%) \\
\hline\hline
a) without aggregation cell & 69.9 \\
b) random aggregation cell & 71.4  \\
c) AutoRTNet-A & 72.9  \\
\hline
\end{tabular}}
% }
\end{center}
\label{aggregationcell}
\end{table}

\subsubsection{Hyper-Cell Searching Process}

\begin{figure}[h]
\centering
\includegraphics[width=7.8cm]{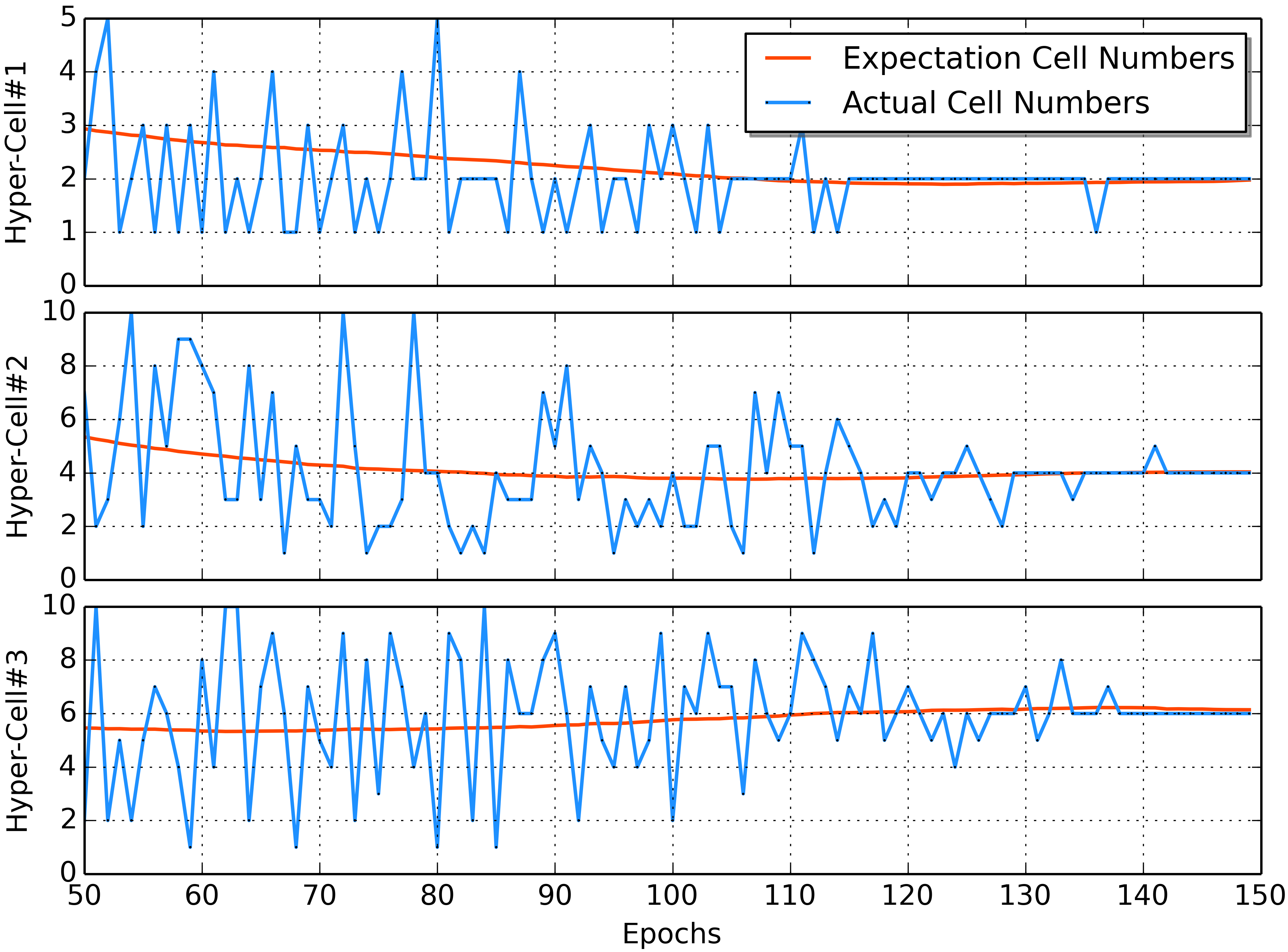}
\caption{Illustration of cell numbers in hyper-cells during the search process. Blue lines from top to bottom denote the actual cell number changing in each hyper-cell with the increase of epochs, and red curves represent the mathematical expectation values of the current cell numbers in hyper-cells.}
\label{figure_bh}
\end{figure}

To better analyze how hyper-cell works throughout the whole searching process, we visualize the number of cells of each hyper-cell after the warm-up phase, as depicted in Figure \ref{figure_bh}. The initial cell numbers are $\left\{ 5, 10, 10 \right\}$ and eventually converges to $\left\{ 2, 4, 6 \right\}$ in three hyper-cells. The blue lines from top to bottom denote the actual cell numbers according to current architecture parameter $\beta$ of each hyper-cell, and red curves represent the mathematical expectation values of current cell numbers. We observe that the framework actively explores different cell numbers (\ie different depths) in each hyper-cell at the early stage of searching, and the expectation values of cell numbers also change gradually. The cell numbers progressively become stable towards the final architecture in the late stage of searching, and the actual cell number lines gradually get close to the expectation curves. Another interesting observation is that hyper-cell \#1 finds its optimal depth much earlier than the other ones, indicating the search process follows a shallow-to-deep manner as we expected.

\subsubsection{Different Latency Settings}

\begin{figure}[h]
\centering
\includegraphics[width=6.5cm]{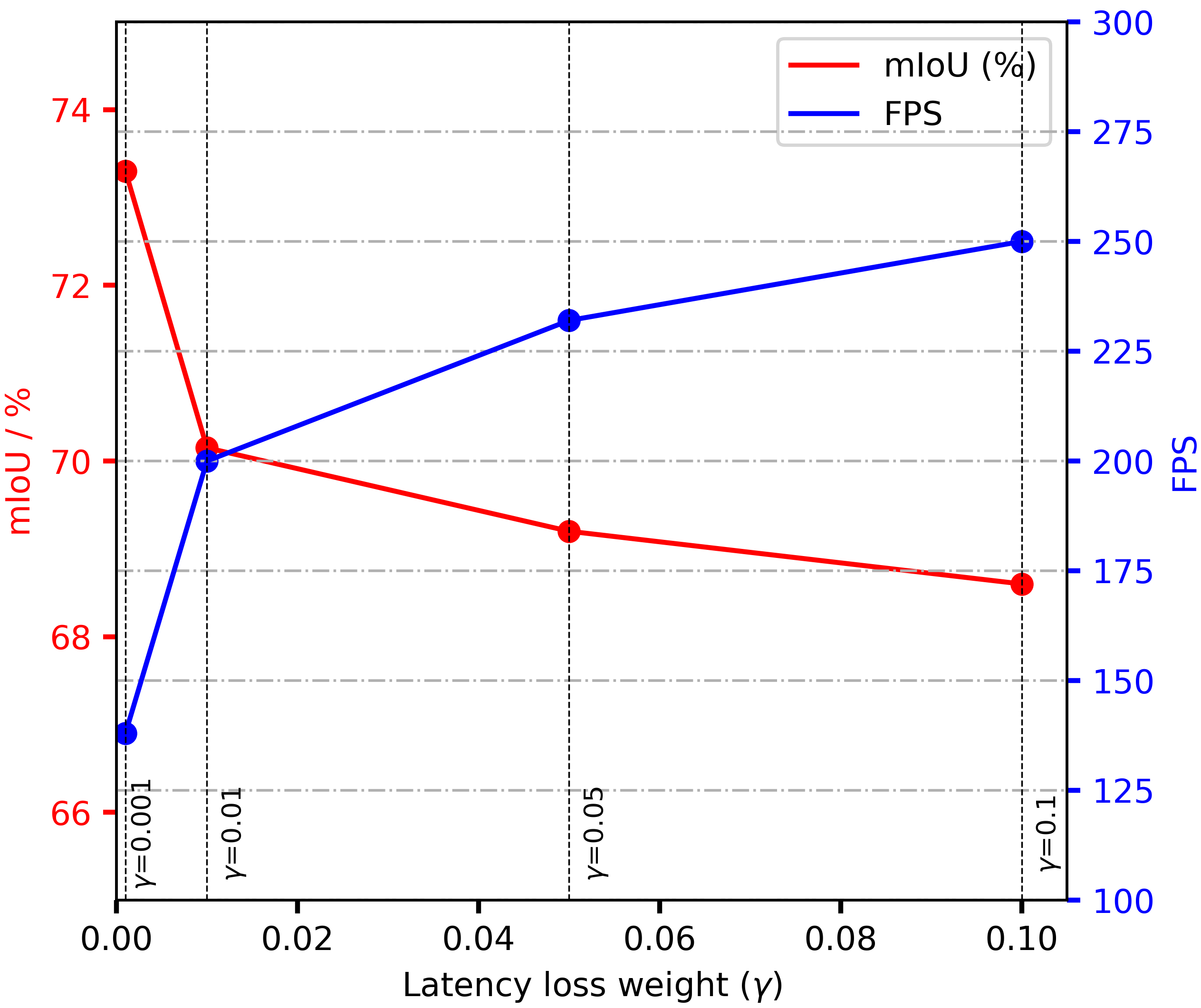}
\caption{The results of different latency settings on CamVid dataset.}
\label{camvid_latency}
\end{figure}

Our joint search framework searches for the optimal network architectures under different latency settings (\ie loss weight $\gamma$). In Section 4.3, AutoRTNet A and B are searched with $\gamma = 0.01$ and $0.001$ on the Cityscapes dataset, respectively, which demonstrates the flexibility of our framework. We also conduct the architecture search on the CamVid dataset with different latency settings, and the results are as depicted in Figure \ref{camvid_latency}. The networks searched with $\gamma$ = 0.001, 0.01, 0.05, 0.1 achieve 73.3\%, 70.2\%, 69.2\%, 68.6\% mIoU and 138.0, 200.2, 232.1, 250.0 FPS on the CamVid test set, respectively. Notably, our AutoRTNet achieves 250.0 FPS while maintaining 68.6\% mIoU, which surpasses ICNet (67.1\% mIoU with 34.5 FPS) and DFANet (64.7\% mIoU with 120 FPS) significantly. 

\subsubsection{Insights from Searched AutoRTNet}

\begin{figure*}
\begin{center}
\includegraphics[width=6.5in]{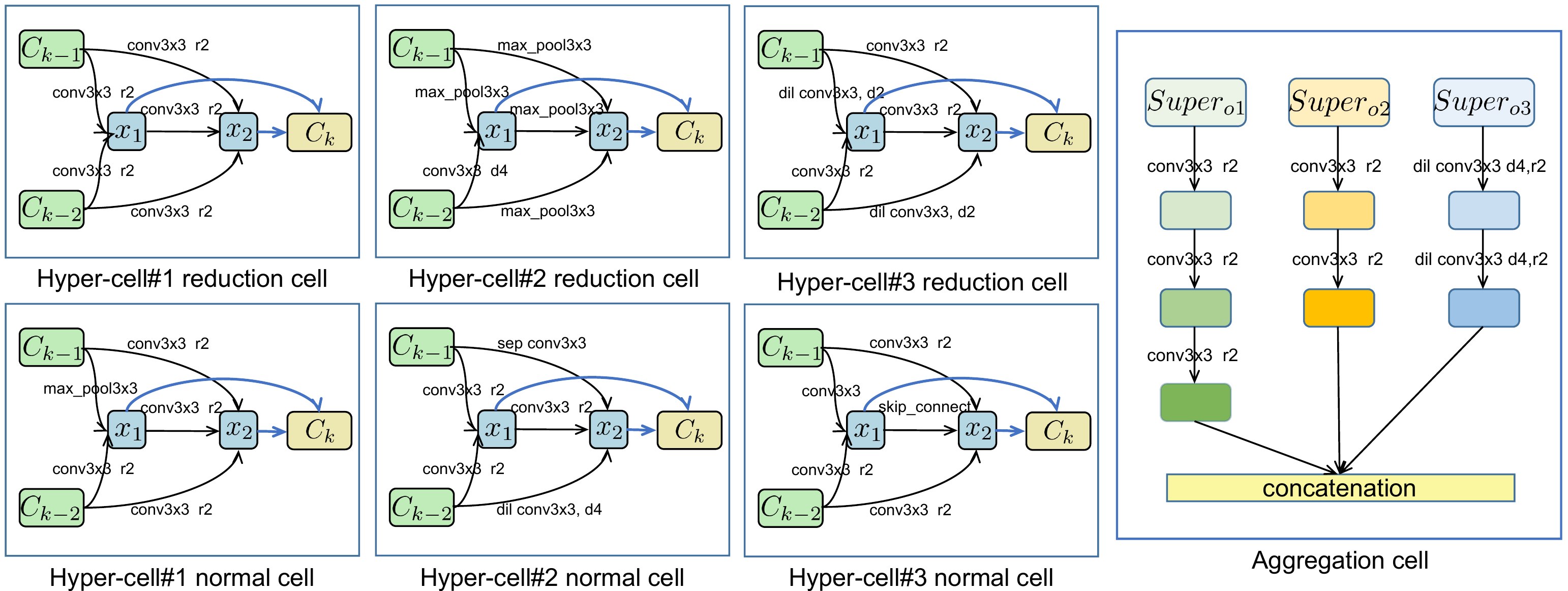}
\end{center}
 \caption{Illustration of the detailed AutoRTNet-A architecture. The structures of the reduction cells and normal cells in three hyper-cells are shown in the figure respectively. The structure of searched aggregation cell in shown on the right. \textbf{Best viewed in color.}}
\label{networkshow}
\end{figure*}

Finally, we provide an in-depth analysis of the AutoRTNet-A searched by our framework. We use the NAS methods to search the suitable architectures for specific tasks, likewise, we should understand why the searched network works well and it will guide the hand-designed process in turn. We have the following three important observations.

\paragraph{Early Downsampling} We notice that the searched downsampling strategy is stable and reasonable. As shown in Table \ref{supercella}, in the first hyper-cell, whether the initialized cell number is 5 or 10, the final number is at most 2 after the optimization. The reason is that the visual information is highly spatially redundant, thus can be compressed into a more efficient representation. Under the latency constraints, the searched downsampling strategy is as we expected and follows the early downsampling \cite{paszke2016enet} priori knowledge.

\paragraph{Suitable receptive field} The suitable receptive field size \cite{luo2016understandingreceptivefield} is crucial for semantic segmentation. Too large receptive field may introduce some extra noise or negative interference, and the network lacks of capturing enough context information if it is too small. During the search process, the AutoRTNet continuously adjusts the operation selection to determine the final receptive field, for example, in Figure \ref{networkshow}, in the optimized aggregation cell, the operations from the outputs of third hyper-cell always choose the operation with dil=4 rather than dil=2 or 8 also in the search space. So we should choose right operations for suitable receptive field in hand-designed real-time semantic segmentation networks.

\paragraph{Operation selection} The early operations act as good feature extractors, as shown in Figure \ref{networkshow}, the selection of operations in early stage always tends to conv3$\times$3. The middle and deep layers have the diversity of operation selection. When performing multi-scale feature aggregation in aggregation cell, as shown in Figure \ref{networkshow}, we clearly found that the deeper layers enjoy dilated convolution, while the lower layers only prefer the common convolution operations.

\subsection{Detailed Time and GPU Information for Fair Comparison}

The inference time or FPS is influenced by the GPU device and the input image size of the model. Here we list detailed information of previous approaches in Table \ref{Cityscapes_big_table} for readers as reference. Our GPU device is Nvidia TitanXP GPU. For fair comparison, we directly quote the reported remeasured or estimated results on TitanXP of other algorithms in CAS \cite{zhang2019customizable} and SwiftNet \cite{orsic2019defenseSwiftNet} paper. And we remeasure the speed of the methods based on our implementation if the original speed was reported on different GPUs and not mentioned in CAS \cite{zhang2019customizable} and SwiftNet \cite{orsic2019defenseSwiftNet}. Note that our implementations and speed measurements do not use TensorRT optimizations.

About the speed gap between the original DFANet \cite{li2019dfanet} and our measurement: The speed of DFANet is reported on TitanX GPU, and also not mentioned in CAS \cite{zhang2019customizable} and SwiftNet \cite{orsic2019defenseSwiftNet}. So we carefully remeasure the inference time on TitanXP for fair comparison. There still has a speed gap between the original speed and the one we measured, we suspect that this is caused by the inconsistency of the implementation platform. We reimplement the DFANet using official PyTorch \cite{paszke2017automaticpytorch}, and they measure it on their own framework in which the depth-wise separable convolution is more fully optimized.

%%% cityscapes
\begin{table*}[h]
\caption{The detailed information of our AutoRTNet and other state-of-the-art methods on the Cityscapes test set. Methods trained using both fine and coarse data are marked with $*$. The mark ${\dag}$ represents the speed is remeasured by us.}
\begin{center}
\setlength{\tabcolsep}{1mm}{
\begin{tabular}{|l|c|c|c|c|c|c|c|}
\hline
\multirow{2}{*}{\textbf{Method}}  & \multirow{2}{*}{\textbf{Input Size }} & \multicolumn{2}{c|}{\textbf{mIoU (\%)}} & \multirow{2}{*}{\textbf{Latency (ms) on TitanXP}} & \multirow{2}{*}{\textbf{FPS on TitanXP}} & \multicolumn{2}{c|}{\textbf{Original Results}} \\
\cline{3-4} \cline{7-8}
& & \multicolumn{1}{c|}{\textit{val}}  & \multicolumn{1}{c|}{\textit{test}} & & & \multicolumn{1}{c|}{\textit{FPS}}  & \multicolumn{1}{c|}{\textit{GPU}}   \\ \hline
\hline
FCN-8S \cite{long2015fullyfcn} & 512x1024  & - & 65.3 & 227.23 & 4.4 &  - & - \\
PSPNet \cite{zhao2017pyramidpspnet}  & 713x713  & - & 81.2 & 1288.0 & 0.78  & - & -  \\
DeepLabV3$^*$ \cite{chen2017rethinkingdeeplabv3} & 769x769 & -  & 81.3 & 769.23 & 1.3 & - & -   \\
AutoDeepLab$^*$ \cite{liu2019autodeeplab} & 769x769 & -  & 81.2 & 303.0 & 3.3 & - & -   \\

\hline
SegNet \cite{badrinarayanan2017segnet} & 640x320  & - & 57.0 & 30.3   & 33    & - & - \\
SQ \cite{treml2016speedingSQ} & 1024x2048 & - &  59.8 & 46.0   & 21.7 & -&Titan X M   \\
ENet \cite{paszke2016enet} & 640x320   & - & 58.3 & 12.7  & 78.4 & 135.4 & Titan X \\
ERFNet \cite{romera2017erfnet}  & 1024x512 & -  & 69.7 & 48.5  & 20.6 &  11.2 & TitanX M \\
ICNet \cite{zhao2018icnet}  & 1024x2048 & 67.7 & 69.5 & 26.5  & 37.7 &  30.3 & TITAN X(M)  \\
SwiftNet \cite{orsic2019defenseSwiftNet} & 1024x2048 & 74.4 &  75.1 & 26.2 & 38.1 & 34.0 & GTX 1080Ti \\
DF1-Seg \cite{li2019partialdfnet} & 768x1536 & 74.1 & 73.0 & 29.1 & 34.4 & 30.7 & GTX 1080Ti \\
ESPNet \cite{mehta2018espnet}  & 1024x512 & - & 60.3 & 8.2 & 121.7 &  112 & TitanX  \\
BiSeNet \cite{yu2018bisenet} & 768x1536 & 69.0 & 68.4 & 9.52   & 105.8 & 105.8 & TitanXP \\
DFANet \cite{li2019dfanet} & 1024x1024 & - & 71.3 & 20.6$^{\dag}$  & 48.5$^{\dag}$ & 100.0 &  TitanX \\
CAS \cite{zhang2019customizable} & 768x1536 & 71.6 & 70.5 & 9.25   & 108.0 & 108.0 &  TitanXP \\
CAS$^*$ \cite{zhang2019customizable}  & 768x1536 & 72.5 & 72.3 & 9.25  & 108.0 & 108.0 &  TitanXP \\
\hline
\textbf{AutoRTNet-A} & 768x1536  & 72.9 & 72.2 & 9.09 & 110.0  & 110.0 & TitanXP  \\
\textbf{AutoRTNet-A$^*$} & 768x1536 & 74.5 & 73.9 & 9.09 & 110.0 & 110.0 & TitanXP \\
\textbf{AutoRTNet-B}  & 768x1536 & 74.7 & 74.3 &  14.0 & 71.4  & 71.4 & TitanXP  \\
\textbf{AutoRTNet-B$^*$}  & 768x1536 & 76.0 & 75.8 & 14.0 & 71.4  & 71.4 & TitanXP \\
\hline
\end{tabular}}
\end{center}
\label{Cityscapes_big_table}
\end{table*}

\subsection{Full Quantitative Results on Cityscapes and CamVid Dataset}

Here we provide detailed quantitative results of per-class mIoU on the Cityscapes and CamVid datasets. Moreover, we provide the performance of the AutoRTNet on the full-resolution Cityscapes validation set.

\subsubsection{Cityscapes Dataset}

Compared with other methods, our AutoRTNet-A achieves overall 72.2\% mIoU with 110.0 FPS, which is the state-of-the-art trade-off between accuracy and speed. The per-class accuracy values are shown in Table \ref{tab:full_city}. In comparison with other methods with public per-class accuracy on the Cityscapes test set, our predictions are more accurate in 13 out of 19 classes. AutoRTNet-A achieves slight improvements on the general classes (Road, Sidewalk, Building, Terrain, Car, etc.), while obtaining a significant accuracy improvement on the challenging classes (Truck, Motorbike, Train, Fence, Rider, etc.). AutoRTNet-B achieves 74.3\% mIoU on Cityscapes test set with 71.4 FPS.

\begin{table*}[h]
\caption{Detailed performance comparison of our AutoRTNet-A with other state-of-the-art methods on the Cityscapes test set.}
\begin{center}
  \resizebox{1.0\textwidth}{!}{
    \begin{tabular}{l|ccccccccccccccccccc|c|c|c}
    \toprule 
    \raisebox{3pt}{Method} & \rotatebox[origin=l]{90}{Road} &
    \rotatebox[origin=l]{90}{Sidewalk} & \rotatebox[origin=l]{90}{Building} &
    \rotatebox[origin=l]{90}{Wall} & \rotatebox[origin=l]{90}{Fence} &
    \rotatebox[origin=l]{90}{Pole} & \rotatebox[origin=l]{90}{Traffic light} &
    \rotatebox[origin=l]{90}{Traffic sign} & \rotatebox[origin=l]{90}{Vegetation } &
    \rotatebox[origin=l]{90}{Terrain} & \rotatebox[origin=l]{90}{Sky}
     &   \rotatebox[origin=l]{90}{Person} & \rotatebox[origin=l]{90}{Rider } &
    \rotatebox[origin=l]{90}{Car} & \rotatebox[origin=l]{90}{Truck}
    &  \rotatebox[origin=l]{90}{Bus} & \rotatebox[origin=l]{90}{Train}
    &  \rotatebox[origin=l]{90}{Motorcycle} & \rotatebox[origin=l]{90}{Bicycle}
    & \rotatebox[origin=l]{90}{Mean IOU(\%) } &
    \rotatebox[origin=l]{90}{FPS} \\
    \midrule
SegNet  &96.4 &73.2 &84.0 &28.4 &29.0 &35.7 &39.8 &45.1 &87.0 &63.8 &91.8 &62.8 &42.8 &89.3 &38.1 &43.1 &44.1 &35.8 &51.9 &57.0 & 33 \\
ENet     &96.3 &74.2 &75.0 &32.2 &33.2 &43.4 &34.1 &44.0 &88.6 &61.4 &90.6 &65.5 &38.4 &90.6 &36.9 &50.5 &48.1 &38.8 &55.4 &58.3 & 78.4 \\
ICNet    &97.1 &79.2 &89.7 &43.2 &48.9 & \textbf{61.5} & \textbf{60.4} &63.4 &91.5 &68.3 &93.5 &74.6 &56.1 & 92.6 &51.3 & 72.7 &51.3 & 53.6 &70.5 &69.5 & 37.7 \\
ESPNet  &97.0 &77.5 &76.2 &35.0 &36.1 &45.0 &35.6 &46.3 &90.8 &63.2 &92.6 &67.0 &40.9 &92.3 &38.1 &52.5 &50.1 &41.8 &57.2 &60.3 & 121.7  \\
ERFNet &97.9 & 82.1 & 90.7 & 45.2 &50.4 &59.0 & 62.6 & \textbf{68.3} & \textbf{91.9} & 69.4 & \textbf{94.2} & \textbf{78.5} & 59.8 & 93.4 & 52.3 & 60.8 & 53.7 & 49.9 & 64.2 & 69.7 & 20.6 \\
BiSeNet & - & - & - & - & - & - & - & - & - & - & - & - & - & - & - & - & - & - & - & 68.4 & 105.8 \\
DFANet & - & - & - & - & - & - & - & - & - & - & - & - & - & - & - & - & - & - & - & 71.3 & 83.1 \\
CAS  & - & - & - & - & - & - & - & - & - & - & - & - & - & - & - & - & - & - & - & 70.5 & \textbf{108.0} \\
 \midrule  
 AutoRTNet-A & \textbf{98.5} & \textbf{84.9} & \textbf{91.4} & \textbf{45.9} & \textbf{53.0} & 52.2 & 60.3 & 67.3 & 91.5 & \textbf{70.6}  & 93.9 & 78.2 & \textbf{62.5} & \textbf{95.3} & \textbf{63.7} & \textbf{74.5} & \textbf{63.9} & \textbf{56.8} & \textbf{67.0} & \textbf{72.2} & 110.0 \\
 \midrule
  AutoRTNet-B & 98.4 & 86.6 & 91.2 & 52.2 & 54.9 & 58.5 & 63.7 & 68.4 & 91.5 & 71.5 & 94.9 & 79.3 & 61.4 & 95.4 & 65.9 & 78.0 & 69.8 & 59.6 & 69.8 & 74.3 & 71.4 \\ 
    \bottomrule
    \end{tabular}%
    }
\end{center}
  \label{tab:full_city}%
\end{table*}%

Cityscapes contains high-resolution 1024 $\times$ 2048 images, which make it a big challenge for real-time semantic segmentation. ICNet \cite{zhao2018icnet} focuses on building a practically fast semantic segmentation system with high-resolution image inputs while accomplishing high-quality results. SwiftNet \cite{orsic2019defenseSwiftNet} and CAS \cite{zhang2019customizable} also perform the experiments on Cityscapes with full-resolution image inputs. In this part, we compare with these methods on the Cityscapes validation set and the results are shown in Table \ref{Cityscapes_full_reso}. We refer to the speed scaling factors on different GPUs in SwiftNet \cite{orsic2019defenseSwiftNet} paper and estimate the speed values of ICNet, SwiftNet, CAS on Titan XP GPU.

%%% cityscapes
\begin{table*}[h]
\caption{Accuracy and speed comparison on the Cityscapes validation set with image resolution 1024 $\times$ 2048.}
\begin{center}
\setlength{\tabcolsep}{1mm}{
\begin{tabular}{|l|c|c|c|c|c|c|}
\hline
\multirow{2}{*}{\textbf{Method}}  & \multirow{2}{*}{\textbf{Input Size}} & \multirow{2}{*}{\textbf{mIoU (\%)}} & \multirow{2}{*}{\textbf{FPS on Titan XP}} & \multicolumn{2}{c|}{\textbf{Original Results}}  \\
\cline{5-6}
& & & & \multicolumn{1}{c|}{\textit{FPS}}  & \multicolumn{1}{c|}{\textit{GPU}}   \\
\hline\hline
SQ \cite{treml2016speedingSQ} & 1024x2048 & 59.8 & -  & - & Titan X(M)  \\
ICNet \cite{zhao2018icnet}  & 1024x2048 & 69.5 & 55.6  & 30.3 &  Titan X(M) \\
SwiftNet \cite{orsic2019defenseSwiftNet} & 1024x2048  &  74.4 & 38.1 & 34.0 & GTX 1080Ti \\
CAS \cite{zhang2019customizable} & 1024x2048  & 74.0 & 45.2  & 34.2 &  GTX 1070 \\
\hline
\textbf{AutoRTNet-A} & 1024x2048 & \textbf{75.0} &  62.7 & 62.7 & Titan XP \\
\textbf{AutoRTNet-B} & 1024x2048 & \textbf{76.8} &  45.6 & 45.6 & Titan XP \\
\hline
\end{tabular}}
\end{center}
\label{Cityscapes_full_reso}
\end{table*}

\paragraph{AutoRTNet-A} Our AutoRTNet-A achieves 75.0\% mIoU and delieves 62.7 FPS on full-resolution Cityscapes val set (\ie 1024 $\times$ 2048). To the best of our knowledge, the real-time performance of AutoRTNet-A outperforms all existing real-time methods. Compared with ICNet, AutoRTNet surpasses it by 5.5\% in mIoU with a faster inference speed. Moreover, AutoRTNet outperforms SwiftNet and CAS by 0.6\% and 1.0\% in mIoU, and has a great advantage in inference speed (\ie 62.7 FPS vs 38.1 FPS, 62.7 FPS vs 45.2 FPS).

\paragraph{AutoRTNet-B} Our AutoRTNet-B achieves 76.8\% mIoU with 45.6 FPS on full-resolution Cityscapes val set, which is the state-of-the-art real-time performance. Compared with SwiftNet and CAS which have a little bit slower speed than us, our AutoRTNet-B surpasses them by 2.4\% and 2.8\% in mIoU, respectively.

\subsubsection{CamVid Dataset}

\begin{table*}[h]
\caption{Detailed performance comparison of our AutoRTNet-A with other state-of-the-art methods on the CamVid test set.}
  \begin{center}
  \resizebox{1.0\textwidth}{!}{
    \begin{tabular}{l|ccccccccccc|c|c|c}
    \toprule 
    \raisebox{3pt}{Method} & \rotatebox[origin=l]{90}{Building} &
    \rotatebox[origin=l]{90}{Tree} & \rotatebox[origin=l]{90}{Sky} &
    \rotatebox[origin=l]{90}{Car} & \rotatebox[origin=l]{90}{Sign} &
    \rotatebox[origin=l]{90}{Road} & \rotatebox[origin=l]{90}{Pedestrian} &
    \rotatebox[origin=l]{90}{Fence} & \rotatebox[origin=l]{90}{Pole } &
    \rotatebox[origin=l]{90}{Sidewalk} & \rotatebox[origin=l]{90}{Bicyclist}
    & \rotatebox[origin=l]{90}{Mean IOU(\%) } &
    \rotatebox[origin=l]{90}{FPS} \\
    \midrule
%  SegNet-Basic & 75.0  & 84.6 &  91.2  & 82.7  & 36.9  & 93.3 & 55.0  & 47.5  & 44.8  & 74.1  & 16.0  & n/a  &  FPS \\
 SegNet   & \textbf{88.8} & 87.3 &  92.4  & 82.1  &  20.5  & 97.2 & 57.1 &  49.3 & 27.5  & 84.4  & 30.7  & 55.6 &  29.4 \\
 ENet     & 74.7  &  77.8  & \textbf{95.1} & 82.4  &  \textbf{51.0} & 95.1 & 67.2  & 51.7  & \textbf{35.4} & 86.7  & 34.1  & 51.3  &  61.2 \\
 ICNet  & -  & - & - & - & - & - & - & - & - & - & - & 67.1 &  34.5 \\
 BiseNet-Xception39   & 82.2  & 74.4  & 91.9  & 80.8  & 42.8  & 93.3  & 53.8 & 49.7 & 25.4  & 77.3 &  50.0 & 65.6  &  - \\
 BiseNet-Res18   &83.0  & 75.8 & 92.0  & 83.7  & 46.5 & 94.6 & 58.8  & 53.6  & 31.9  &  81.4  & 54.0  & 68.7  &  - \\
 DFANet  & -  & - & - & - & - & - & - & - & - & - & - & 64.7 &  120.0 \\
 CAS  & -  & - & - & - & - & - & - & - & - & - & - & 71.2 & 169.0 \\
 \midrule
AutoRTNet-A & 88.1  &  \textbf{78.4}  &  91.7  &  \textbf{93.0}  & 44.0 &  \textbf{96.2}  & \textbf{66.8} &  \textbf{60.8} & 33.0 & \textbf{88.6} &  \textbf{67.4}  & \textbf{73.5} &  140.0  \\
\midrule
AutoRTNet-B & 89.1  & 79.3 &  91.7  &  92.5  & 41.0 & 96.8 & 64.7 & 66.5 & 32.2 & 90.0 &  72.1  & 74.2 &  82.5  \\
AutoRTNet-C & 87.5  &  77.9  &  91.7  &  90.6 & 30.4 &  95.6  & 62.2 &  47.1 & 25.2 & 87.5 & 58.9 & 68.6 & \textbf{250.0}  \\
    \bottomrule
    \end{tabular}%
    }
\end{center}
  \label{tab:full_camvid}%
\end{table*}%

As shown in Table \ref{tab:full_camvid}, with 720 $\times$ 960 input images, the searched AutoRTNet-A achieves 73.5\% mIoU with 140.0 FPS, which is the state-of-the-art trade-off between accuracy and speed on the CamVid test set. In comparison with other methods, the predictions of our AutoRTNet-A are more accurate in 7 out of the 11 classes. More importantly, the inference speed of AutoRTNet-A achieves 140 FPS, which is very impressive compared with other methods. (\eg SegNet 29.4 FPS, ENet 61.2 FPS, ICNet 34.5 FPS). The per-class accuracy of AutoRTNet B and C are also shown in Table \ref{tab:full_camvid}.

\subsubsection{Visual Segmentation Results}

\begin{figure*}[t]
\centering
\includegraphics[width=18cm]{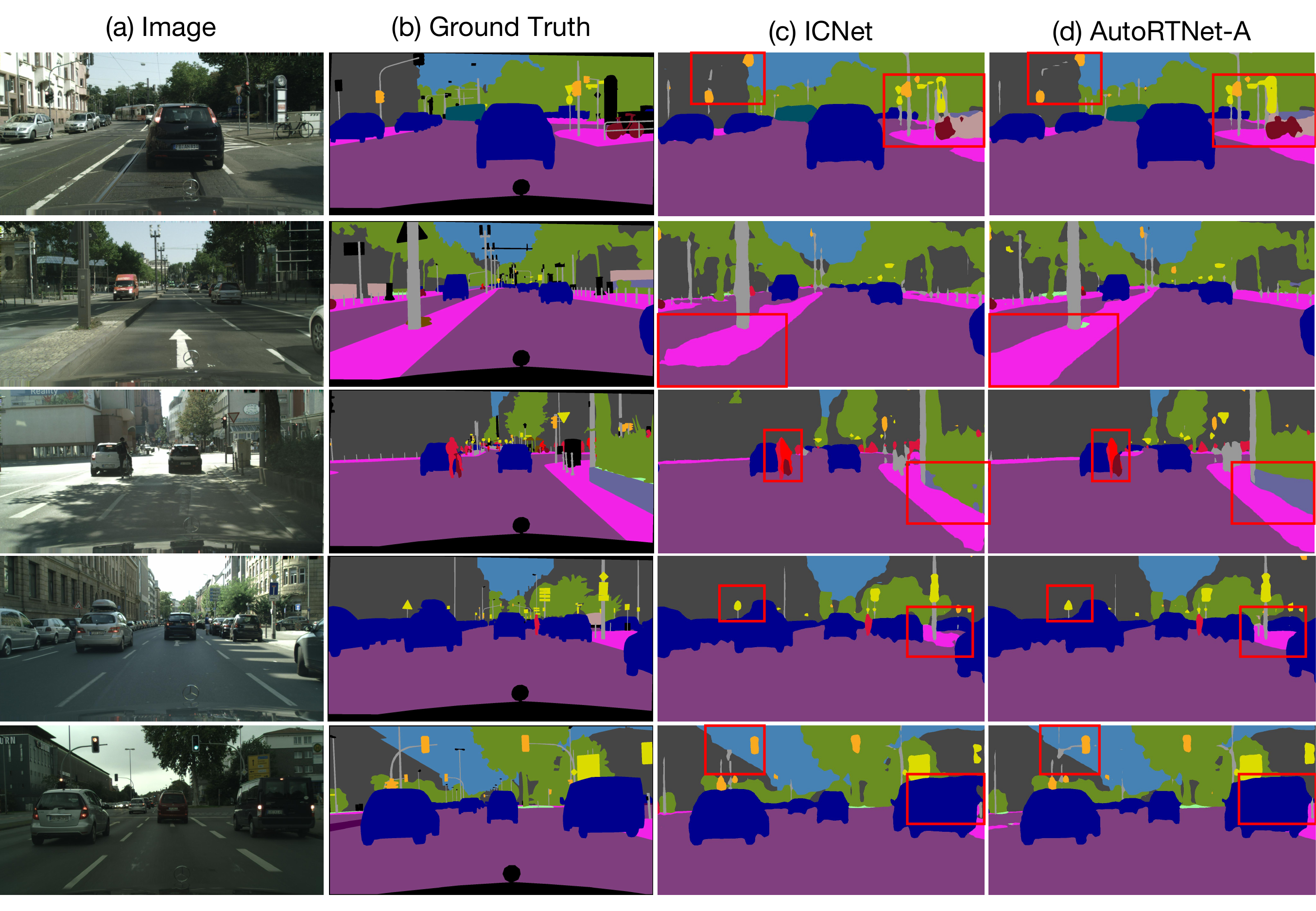}
\caption{Visual segmentation results on Cityscapes validation set. (a) Image. (b) Ground Truth. (c) ICNet. (d) AutoRTNet-A.}
\label{cityscapes_show}
\end{figure*}

\begin{figure*}[h]
\centering
\includegraphics[width=18cm]{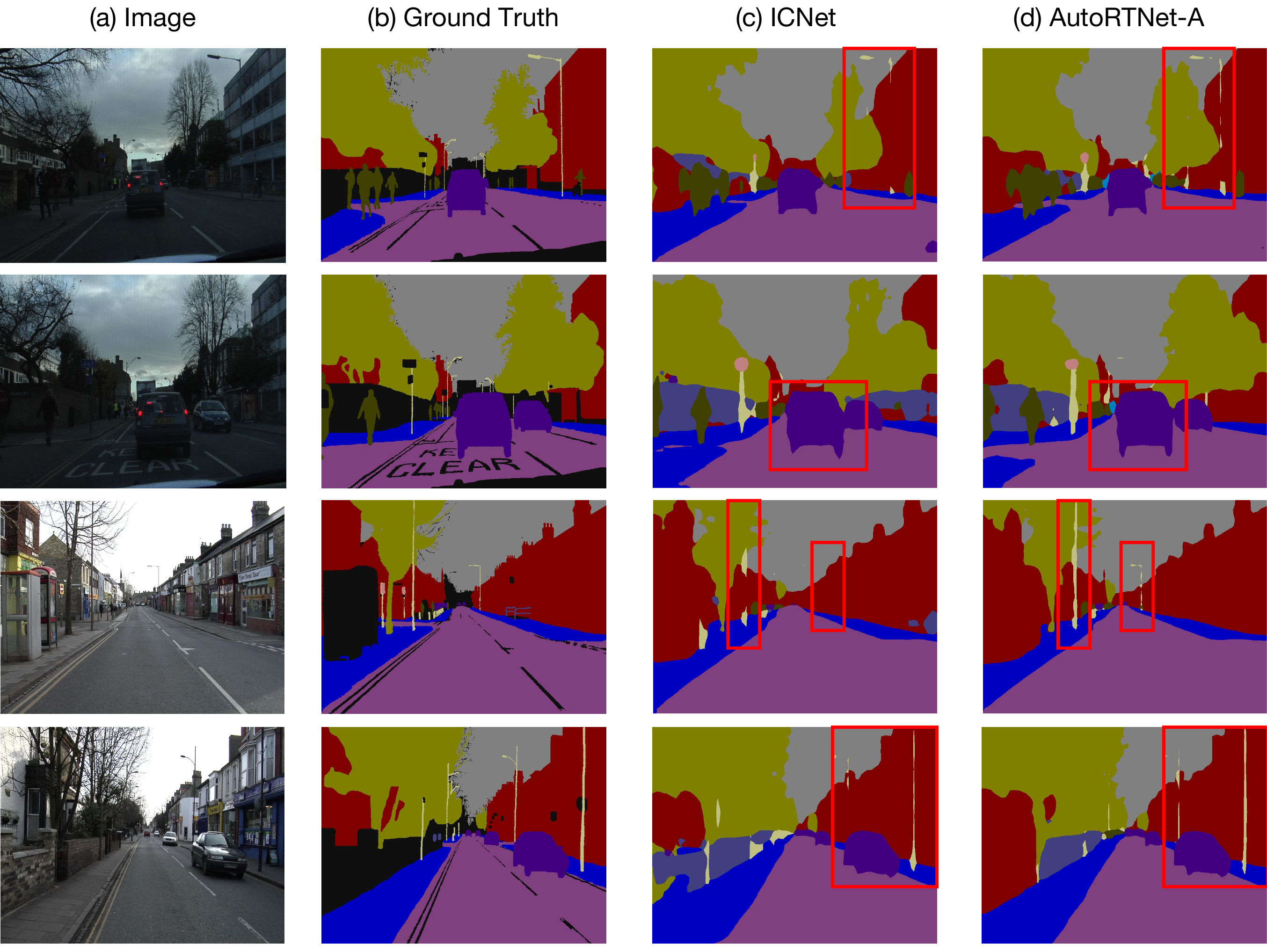}
\caption{Visual segmentation results on CamVid test set. (a) Image. (b) Ground Truth. (c) ICNet. (d) AutoRTNet-A. }
\label{camvid_show}
\end{figure*}

We provide some visual prediction results on both Cityscapes and CamVid datasets here. As shown in Figure \ref{cityscapes_show} and Figure \ref{camvid_show}, the columns correspond to input image, ground truth, the prediction of ICNet, and the prediction of our AutoRTNet-A. Compared with ICNet, AutoRTNet-A produces more accurate and detailed results with faster inference speed. For example, AutoRTNet-A captures small objects in more details (\eg traffic light in Figure \ref{cityscapes_show}, poles in Figure \ref{camvid_show}) and generates ``smoother'' results on object boundaries (\eg rider, fence in Figure \ref{cityscapes_show}, car in Figure \ref{camvid_show}). 

\section{Conclusion}

In this paper, we propose a novel joint search framework which covers all three main aspects of the design philosophy for real-time semantic segmentation networks. The framework searches for building blocks, network depth, downsampling strategy, and feature aggregation way simultaneously. The hyper-cell is proposed for searching for the network depth and downsampling strategy in an adaptive manner, and the aggregation cell is introduced for automatic multi-scale feature aggregation. Extensive experiments on both Cityscapes and CamVid datasets demonstrate the superiority and effectiveness of our approach.

\bibliographystyle{ieee_fullname}
\bibliography{egbib}

\end{document}